\documentclass{article}

\usepackage{microtype}
\usepackage[]{graphics}
\usepackage{subfigure}
\usepackage{booktabs} % for professional tables
\usepackage{xcolor, soul}

\usepackage{hyperref}

\usepackage[accepted]{icml2023}

\usepackage{amsmath}
\usepackage{amssymb}
\usepackage{mathtools}
\usepackage{amsthm}

\usepackage[capitalize,noabbrev]{cleveref}

\theoremstyle{plain}

\theoremstyle{definition}

\theoremstyle{remark}

\usepackage[textsize=tiny]{todonotes}

\usepackage{amsfonts}
\usepackage{amsmath}
\usepackage{amsthm}
\usepackage{xspace}
\usepackage{amsmath,amssymb}
\usepackage{hyperref}
\usepackage{bm}

\hypersetup{
    colorlinks=true,
    linkcolor=blue,
    filecolor=magenta,      
    urlcolor=cyan,
    pdfpagemode=FullScreen,
    citecolor=magenta
    }

\usepackage{ifthen}

\usepackage{xcolor}
\usepackage{amsmath,amssymb,amsthm}
\usepackage[inline]{enumitem}

\newboolean{showcomments}
\setboolean{showcomments}{true}
\ifthenelse{\boolean{showcomments}}
{

}

\newcommand{\V}{\mathcal{V}}

\newcommand{\Adapter}{{\texttt{$k$NN-Adapter}}}

\icmltitlerunning{\Adapter: Efficient Domain Adaptation for Black-Box Language Models}

\begin{document}
\twocolumn[
\icmltitle{\Adapter: Efficient Domain Adaptation for Black-Box Language Models}

\icmlsetsymbol{equal}{*}

\begin{icmlauthorlist}
\icmlauthor{Yangsibo Huang}{princeton}
\icmlauthor{Daogao Liu}{uw}
\icmlauthor{Zexuan Zhong}{princeton}
\icmlauthor{Weijia Shi}{uw}
\icmlauthor{Yin Tat Lee}{uw,msr}
\end{icmlauthorlist}

\icmlaffiliation{princeton}{Princeton University}
\icmlaffiliation{uw}{University of Washington}
\icmlaffiliation{msr}{Microsoft Research}

\icmlcorrespondingauthor{Yangsibo Huang}{yangsibo@princeton.edu}

\icmlkeywords{Machine Learning, ICML}

\vskip 0.3in
]



\printAffiliationsAndNotice{}  % leave blank if no need to mention equal contribution

\setlength{\textfloatsep}{10pt}

\begin{abstract}

Fine-tuning a language model on a new domain is standard practice for domain adaptation. However, it can be infeasible when it comes to modern large-scale language models such as GPT-3, which can only be accessed through APIs, making it difficult to access the internal parameters of the model.
In this paper, we propose \Adapter, a method to effectively adapt these black-box large language models (LLMs) to a new domain. The \Adapter~builds on top of the retrieval-augmented language model, and adaptively learns to interpolate the output of the language model with retrieval results from a datastore consisting of the target domain data. Our experiments on four different domains demonstrate that {\Adapter} significantly improves perplexity, and works particularly well in settings with limited access to LLMs. Additionally, we show that \Adapter~is more effective than fine-tuning when the amount of training data is limited. We also release a dataset to encourage further study.

\end{abstract}
\section{Introduction}

\begin{figure}[t]
% \vskip -0.1in
\begin{center}
\includegraphics[width=1.05\columnwidth]{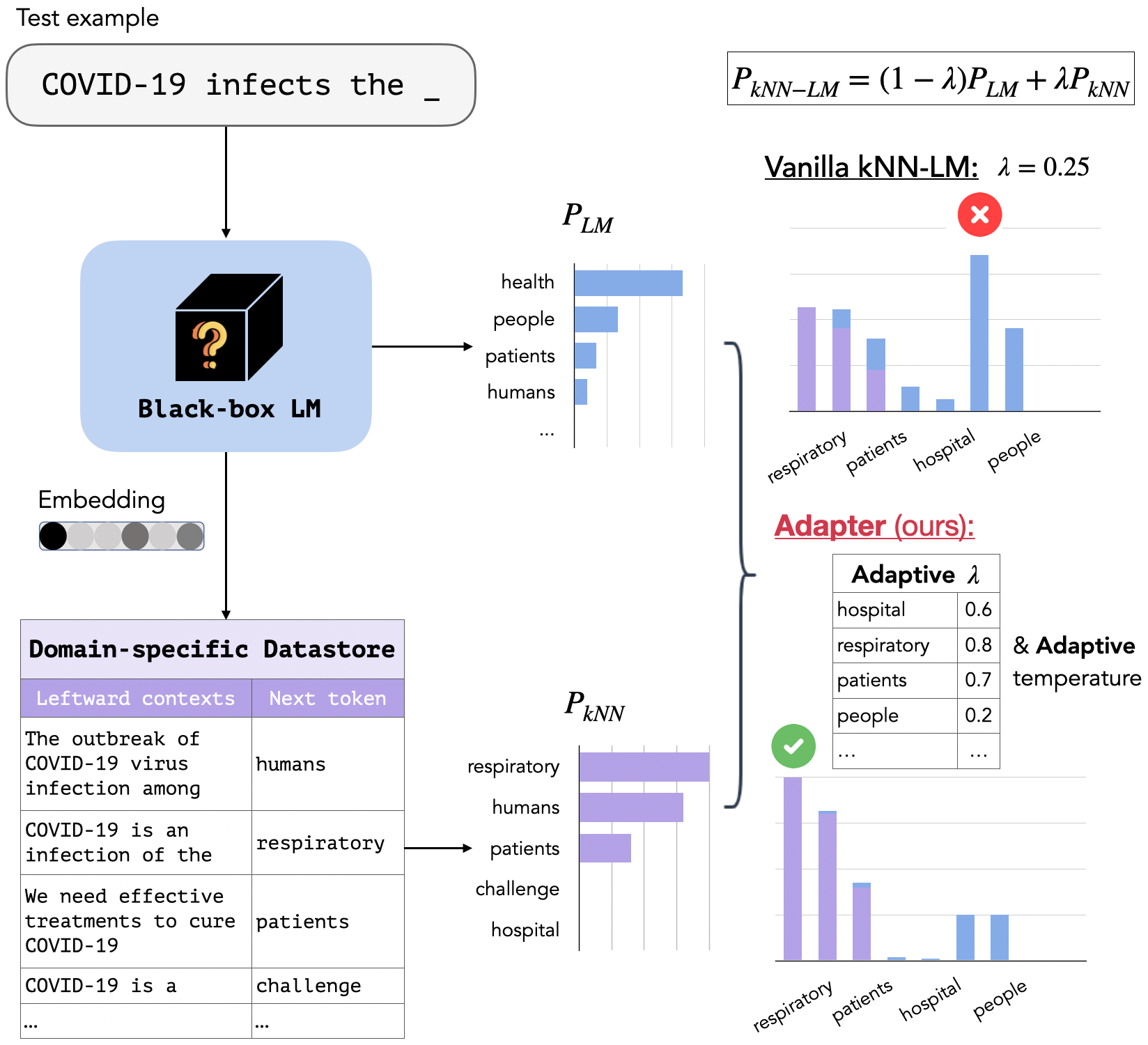}
\end{center}
\vskip -0.1in
\caption{Illustration of \Adapter. Large language models (LLMs) are often deployed as black-box APIs, making it infeasible to fine-tuning them. \Adapter~is an alternative to fine-tuning which effectively adapts a pre-trained LM to a target domain. It is built upon $k$NN-LM~\cite{khandelwal2019generalization} - instead of using a fixed interpolation coefficient $\lambda$, the \Adapter~learns a token-wise $\lambda$ which best combines $p_{\rm LM}$ (from the pretrained LM) and $p_{\rm kNN}$ (from the domain-specific datastore), and thus improves the domain adaptation performance of the original $k$NN-LM.}
\label{main_figure}
\end{figure}

Large language models (LLMs)~\cite{radford2019language, shoeybi2019megatron, brown2020language, raffel2020exploring, zhang2022opt} have achieved state-of-the-art performance in a wide range of natural language processing tasks. However, general-purpose LLMs that are pretrained on one domain (e.g., web crawl data) may not perform well when applied to new domains (e.g., electronic health records), as the common vocabulary, style, and context can be vastly different between domains. 

A standard practice to adapt a pretrained language model to a new domain (i.e., domain adaptation) is model fine-tuning~\cite{gururangan2020don} -- by fine-tuning a language model on a new domain, we can improve its performance on that domain without having to retrain the entire model from scratch. This can be especially useful in cases where there is limited labeled data available for the target domain, or when the target domain has specific characteristics that require a specialized model. Unfortunately, fine-tuning can be infeasible when it comes to modern large-scale LLMs such as GPT-3~\cite{brown2020language}, which are often deployed as \textit{black-box} systems such as Application Programming Interface (API)\footnote{These APIs usually support 1)   \href{https://beta.openai.com/docs/guides/embeddings}{text embedding}, and 2) the \href{https://beta.openai.com/docs/api-reference/completions}{probability of next tokens} given the current context.}.
% , where users cannot access the internal parameters of the model
% The black-box nature makes them infeasible for finetuning for specific use cases. 
Due to the black-box constraint, users cannot access the internal parameters of the model and fine-tune them for their specific use cases.
% . difficult to access the internal parameters of the model, and thus to fine-tune it for specific use cases.
Additionally, fine-tuning may not be a viable option for users with computational constraints, as it requires gradient back-propagation of the entire model which is costly in terms of both computation time and memory (e.g., GPT-3 has 175B parameters).

%\zexuan{The motivation here can probably be made even stronger. Compared to fine-tuning, our approach (1) only requires black-box APIs; (2) doesn't require backward pass to the whole model (which is much more efficient); (3) the model footprint is much smaller (in fine-tuning we need to save 175B new parameters for each domain vs here we only introduce and need to save a small amount of parameters for each domain.)} \yang{I added these points. Please check (and feel free to edit!)}

$k$-Nearest Neighbor Language Model~\cite{khandelwal2019generalization} has been proposed as a way to effectively adapt LLMs to a new domain without fine-tuning. 
% It combines the predictions of a language model with those made by the top-$k$ most similar examples from a datastore built from the target domain.
It combines the output of a language model with the predictions made by the top-$k$ nearest matching examples from a datastore built from the target domain. This combination allows the model to adapt to the new target domain by incorporating the specific characteristics of that domain into its predictions without additional training. However, the zero-shot nature of retrieval-augmented domain adaptation often results in limited utility, as the model is not trained on the target domain, but rather adapts to the domain only based on the nearest examples it can find in the datastore. This can lead to suboptimal performance compared to models that are specifically fine-tuned on the target domain.

To tackle this problem, we propose \Adapter, a lightweight architecture built on top of the $k$-Nearest Neighbor Language Model ($k$NN-LM). \Adapter~improves the domain adaptation performance of $k$NN-LM by learning to adjust two key parameters, the interpolation coefficient, and the distribution temperature, based on the token to be predicted, the current context, and the retrieved neighbors from the datastore (see Figure~\ref{main_figure}).

The core contributions of this paper are:
\begin{enumerate}
    \vspace{-2mm}
    \item We evaluate the performance of \Adapter~on various domains and models with varied sizes (Section~\ref{sec:main_results}). Across all settings, it shows a significant improvement of around 2 perplexities over the $k$-Nearest Neighbor Language Model with a slight extra computational cost. 
    \vspace{-1.5mm}
    \item Additionally, we find that \Adapter~works exceptionally well when the user only has limited access to the language model, for instance, when the API only returns the probability of the top tokens instead of the full probability). \Adapter~is able to maintain high performance even in scenarios where the target domain has very limited data (Section~\ref{sec:limitation}).
    \vspace{-1.5mm}
    \item We also demonstrate that \Adapter~is more effective than directly fine-tuning the LLM on the target domain in the few-shot settings (Section~\ref{sec:compare_w_finetune}), namely when using fewer than $10^4$ training tokens. Furthermore, \Adapter~is much more efficient than fine-tuning the language model, as it has fewer trainable parameters (several thousand) compared to modern LLMs (usually several million or billion).
    \vspace{-1.5mm}
    \item We finally use the adaptively learned parameters by the \Adapter~to gain a better understanding of the source of utility gained by $k$-Nearest Neighbor Language Model (Section~\ref{sec:analysis}).
    \vspace{-1.5mm}
    \item Besides, we plan to release a dataset for researchers to further investigate and improve upon the method proposed in this work. This dataset will be made available with the final version of this paper.
\end{enumerate}

\section{The \Adapter}

In this section, we first describe $k$-Nearest Neighbor Language Model (Section~\ref{sec:knnlm}), and then introduce our \Adapter~(Section~\ref{sec:adapter}).

\subsection{$k$-Nearest Neighbor Language Model}
\label{sec:knnlm}

$k$-Nearest Neighbor Language Model ($k$NN-LM)~\cite{khandelwal2019generalization} is a type of retrieval-augmented language model, in which the model interpolates the prediction from a language model with $k$ nearest responses selected from a pre-defined set of possible contexts (usually referred to as a datastore). 

\vspace{-2mm}
\paragraph{Encoding function.} Given a vocabulary $\mathcal{V}$, the encoding function $f(\cdot)$ maps a sequence $s \in \mathcal{V}^*$ to a fixed-dimensional vector representation of size $d$. Typically, the function $f(\cdot)$ is computed using a language model, where $f(s)$ is the vector representation of the output layer when input $s$ is processed. 

\vspace{-2mm}
\paragraph{Datastore.} 
The datastore is a key-value store generated by
running the encoding function $f(\cdot)$ over a corpus of context.  
Each key, represented as $f(c)$, corresponds to a vector representation of some context $c \in \mathcal{V}^*$, and each value, $v \in \mathcal{V}$, corresponds to the corresponding ground-truth next token for the context $c$.

\vspace{-2mm}
\paragraph{Inference.}  At inference time when predicting the next token for an input sequence $x \in \mathcal{V}^*$, the model queries the datastore with $f(x)$ to retrieve $x$'s $k$-nearest neighbors $\mathcal{N}_k$ according to a distance function $d(\cdot, \cdot)$\footnote{$d(\cdot, \cdot)$ is typically the $\ell_2$ distance in the literature.}. Then the model computes a softmax over the (negative) distances, which gives a distribution over the next token:
\begin{equation}
p_{\mathrm{kNN}}(y|x) \propto \sum_{\left(c_i, w_i\right) \in \mathcal{N}_k} \mathbf{1}_{y=w_i} \exp \left(-\frac{d\left(c_i, f(x)\right)}{t}\right),
\label{eq:normalize_knn}
\end{equation}

where $t$ is a temperature parameter. The prediction is then interpolated with the prediction from the language model:
\begin{equation}
p(y|x) = \lambda p_{\mathrm{kNN}}(y|x) + (1-\lambda) p_{\mathrm{LM}}(y|x),
\end{equation}
where $\lambda$ is an interpolation coefficient.

\subsection{Domain Adaptation with \Adapter}
\label{sec:adapter}

$k$NN-LM can effectively adapt pre-trained language models to new domains by using a datastore built from examples from the target domain~\cite{khandelwal2019generalization}. However, the zero-shot nature of this procedure often leads to limited performance. To overcome this, we propose the use of \Adapter, which adaptively learns to interpolate the output of the pre-trained language model with retrieval results for improved domain adaptation performance.

Our \Adapter~targets at two crucial hyper-parameters in $k$NN-LM:  the interpolation coefficient $\lambda$ and the temperature $t$. Vanilla $k$NN-LM employs both $\lambda$ and $t$ as fixed values for any given query, which may lead to a lack of flexibility and underutilization of the retrieved information. Instead, we propose to \textit{learn} these hyper-parameters in a trainable fashion, as described in the following section.

\begin{figure}[t]
\vskip 0.1in
\begin{center}
\subfigure{\includegraphics[width=0.49\columnwidth]{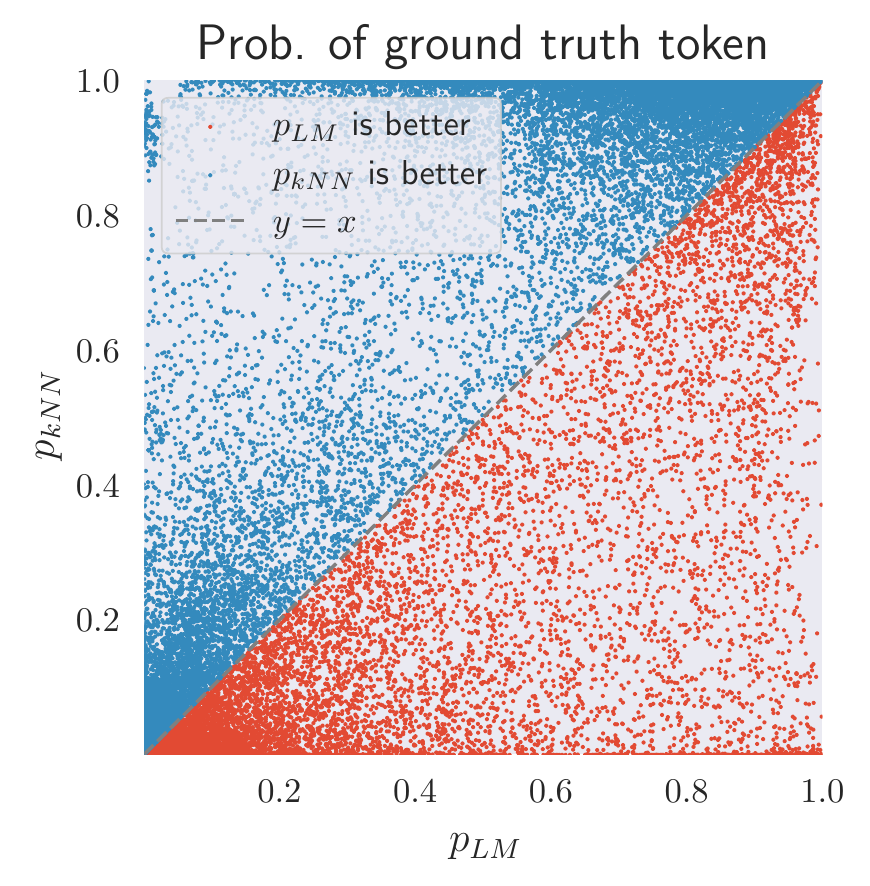}}
\subfigure{\includegraphics[width=0.49\columnwidth]{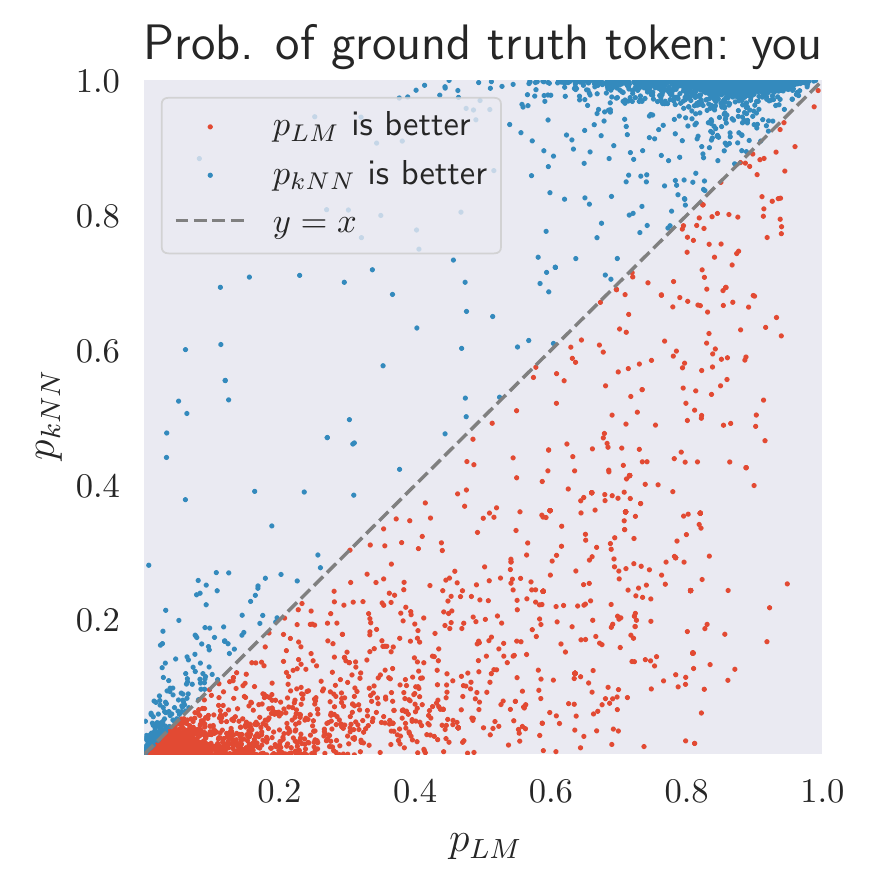}}
\end{center}
\vskip -0.2in
\caption{Probability of the ground-truth tokens in $p_{\rm LM}$ and $p_{\rm kNN}$ for different tokens (left), and the same token (i.e., ``you") but in different sequences (right). }
\label{compare_lm_knn}
\end{figure}

\begin{figure}[t]
\vskip 0.1in
\begin{center}
\includegraphics[width=\columnwidth]{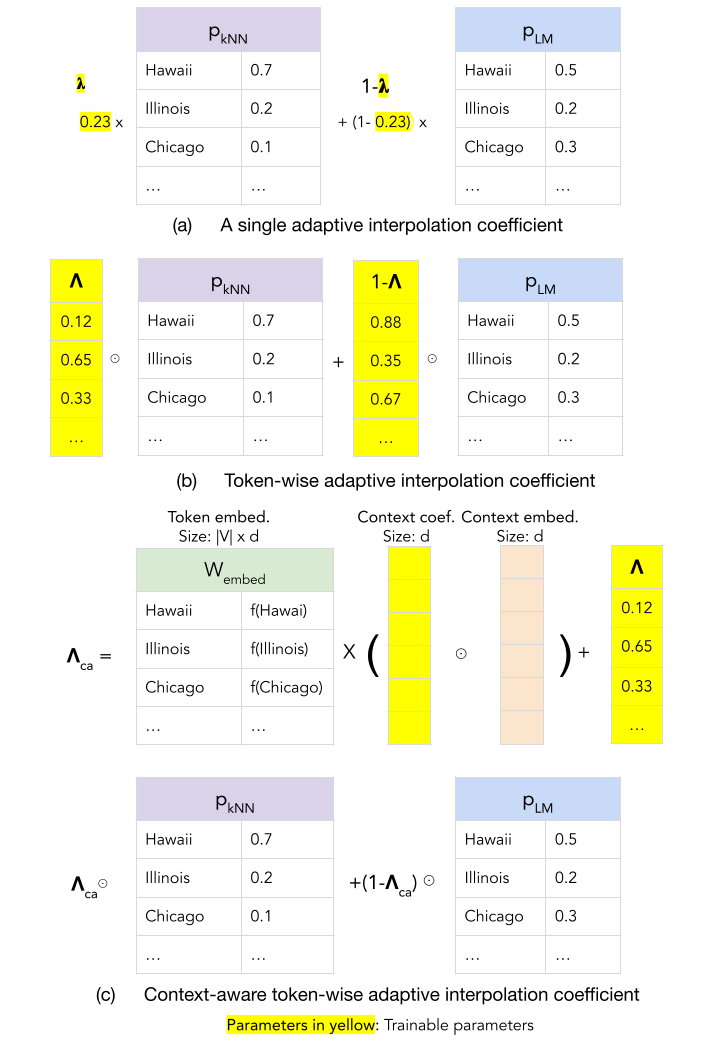}
\vskip -0.1in
\caption{Illustration of three design choices for adaptive interpolation coefficients. Note that after interpolation, we take the logits and compute the softmax on the logits to ensure that the interpolated result is a normalized distribution.
}
\label{arch_figure}
\end{center}
\end{figure}

\subsubsection{Adaptive Interpolation Coefficient}
\label{sec:adapt_lmd}

Let's start with some evidence for why using a fixed value for $\lambda$ may be suboptimal. We observe that the probabilities estimated by $p_{\rm LM}$ and $p_{\rm kNN}$ show different behaviors for different tokens: as shown in the left plot of Figure~\ref{compare_lm_knn}, when being the ground truth prediction, some tokens have a larger probability predicted by $p_{\rm LM}$ (red dots), while others have a larger probability by $p_{\rm kNN}$ (blue dots). It is reasonable to conjecture that LM may be more effective at predicting certain tokens, while the $k$NN  performs better for others. Therefore, an ideal solution is to use a smaller interpolation coefficient ($\lambda$) for the former, and a larger one for the latter. Moreover, as shown in the right plot of Figure~\ref{compare_lm_knn},
 even for the same token, whether LM or $k$NN is better may depend on the concrete context.

Based on these observations, we consider three different design choices for the adaptive interpolation coefficient: single, token-wise, and context-aware token-wise, which are defined as follows.

Fixing the query sentence $x\in \mathcal{V}^*$,
recall that we find the $k$-nearest neighbors ${\cal N}_k=\{(c_1,\omega_1),(c_2,\omega_2),\cdots,(c_k,\omega_k)\}$, which without loss of generality is sorted ascending by the distance, i.e., $d(c_i,f(x))\le d(c_j,f(x))$ if $i<j$, and get the prediction from the original LM $p_{\mathrm{LM}}(y|x)$. 

\paragraph{A single adaptive $\lambda$.} Single adaptive interpolation coefficient (i.e.,  $\lambda$.) is a function that maps ${\cal N}_k$ and $p_{\mathrm{LM}}(y|x)$ to a scalar value in $(0,1)$, which is denoted by $\lambda\big({\cal N}_k,p_{\mathrm{LM}}(y|x)\big)$.
The output prediction should be
\begin{align}
    p(y\mid x) &= \lambda\big({\cal N}_k,p_{\mathrm{LM}}(y|x)\big)  p_{\mathrm{kNN}}(y|x)\nonumber\\
    &+\left(1-\lambda\big({\cal N}_k,p_{\mathrm{LM}}(y|x)\big) \right) p_{\mathrm{LM}}(y|x),
\end{align}

% \vspace{-2mm}
\paragraph{Token-wise adaptive $\Lambda$.} The token-wise adaptive interpolation coefficient maps ${\cal N}_k$ and $p_{\mathrm{LM}}(y|x)$ to a vector $(0,1)^{|\mathcal{V}|}$ instead of a single scalar, which we denote by $\Lambda\big({\cal N}_k, p_{\mathrm{LM}}(y|x)\big)$.
The final prediction is interpolated as:
\begin{align}
    p(y\mid x) &\propto \Lambda\big({\cal N}_k, p_{\mathrm{LM}}(y|x)\big) \odot  p_{\mathrm{kNN}}(y|x)\nonumber\\
    &+ \big(\mathbf{1}-\Lambda\big({\cal N}_k, p_{\mathrm{LM}}(y|x)\big)\big)\odot p_{\mathrm{LM}}(y|x),
\end{align}
where $\mathbf{1}$ is a $|\V|$-dimensional all-one vector, and $\odot$ is the Hadamard product. 

\paragraph{Context-aware token-wise adaptive $\Lambda_{ca}$.} The last two coefficients do not use the information of the query sentence $x$ and hence are context-free. 
We also propose to use the context information to further adjust the token-wise $\Lambda$.

We first collect $W = \{f(v)\}_{v\in{\cal V}} \in \mathbb{R}^{|\V| \times d}$, the  embedding for each token in the vocabulary, where $d$ is the embedding size. Note that $W$ is a fixed matrix as $f(\cdot)$ is never updated during the training of \Adapter. We then combine the token and context information and obtain
\begin{align}
    \Lambda'(x,{\cal N}_k,p_{\mathrm{LM}}(y|x)) = W (\Sigma\big({\cal N}_k,p_{\mathrm{LM}}(y|x)\big)\odot f(x)),
\end{align}
where $\Sigma\big({\cal N}_k,p_{\mathrm{LM}}(y|x)\big) \in \mathbb{R}^d$  is a trainable vector that controls the signal strength for each embedding entry. 

Finally, we add up\footnote{The motivation for the summation is that $\Lambda$ already captures an appropriate interpolation coefficient for each token, and by adjusting its value slightly (up or down) based on the context, we can further improve the  performance.} the context-aware $\Lambda'$ with the token-wise $\Lambda$ to obtain the context-aware token-wise interpolation coefficient (see Figure~\ref{arch_figure}(c)),
\begin{align}
    \Lambda_{ca} = \Lambda({\cal N}_k,p_{\mathrm{LM}}(y|x)) + \Lambda'({\cal N}_k,p_{\mathrm{LM}}(y|x),f(x)).
\end{align}

The distribution of the next word is
\begin{align}
    p(y\mid x) &\propto \Lambda_{ca}\odot p_{\mathrm{kNN}}(y|x)\nonumber\\
    &+\big(\mathbf{1}-\Lambda_{ca}\big)\odot p_{\mathrm{LM}}(y|x).
\end{align}

\subsubsection{Adaptive Temperature}
\label{sec:adapt_temp}

The temperature parameter in $k$-nearest neighbor interpolation can also impact the performance of the model. As the temperature increases, the kNN model gives more equal weight to the output of all nearest neighbors, whereas a decrease in temperature causes the model to give more weight to the closest neighbors (see Equation~\ref{eq:normalize_knn}). We propose two ways to adaptively learn the temperature. 

\paragraph{A single adaptive $t$.} The single adaptive temperature $t\in\mathbb{R}$ is a  trainable scalar that can be updated to optimize the $k$NN distribution. It is used in the same way as a fixed temperature value in a standard $k$NN-LM during the inference step.

\paragraph{Neighbor-wise adaptive $T$.} The neighbor-wise adaptive temperature $T\in\mathbb{R}^k$ is a trainable $k$-dimensional vector, which learns a different temperature for each neighbor. With the neighbor-wise adaptive temperature, the distribution over the next token is computed as follows:
\begin{align}
    p_{\mathrm{kNN}}(y\mid x,T)\propto \sum_{i\le k}\mathbf{1}_{y=\omega_i}\exp\left(-\frac{d(c_i,f(x))}{T_i}\right).
\end{align}
\begin{table*}[t]
\setlength{\tabcolsep}{2.5pt}
\vskip 0.05in
\begin{center}
\begin{small}
\begin{sc}
\begin{tabular}{lcccccc}
\toprule
 {\bf Model} & $\#$ {\bf Extra Trainable Param.} & {\bf WikiText-103} & {\bf BookCorpus} &  {\bf Enron} & {\bf Amazon Reviews } \\
\midrule
{\bf Standard LM }    & 0 & 35.92 & 20.22 &   18.94 & 24.61 \\
\midrule
{\bf $k$NN-LM}   &  0 & \underline{30.77}  & \underline{12.49} &  \underline{13.09}  & \underline{14.02}\\
+ a single adaptive $t$  & $1$ & 30.64 & 12.47 & 13.08 & 13.97\\
+ neighbor-wise adaptive $T$ & $k$ & 30.61 & 12.47 & 13.08 & 13.96  \\ 
\midrule
{\bf $k$NN-LM w/ a single adaptive $\lambda$} & $1$ & 30.63 & 12.47 & 12.98 & 14.07 \\
+ a single adaptive $t$     &  $2$ & 30.56 & 12.45 & 12.97 & 14.05 \\
+ neighbor-wise adaptive $T$     &  $1+k$ & 30.55 & 12.44 & 12.97 & 14.08 \\
\midrule
{\bf $k$NN-LM w/ adaptive token-wise $\Lambda$}    & $| \V |$ &  28.26 & 10.37 &  11.79  & 12.88\\
+ a single adaptive $t$     &  $| \V | + 1$  & 28.20 & \textbf{10.35} & 11.79&  \textbf{12.85}\\
+ neighbor-wise adaptive $T$ & $| \V | + k$       & \textbf{28.17} & \textbf{10.35} & 11.77 & 12.89 \\
\midrule
{\bf $k$NN-LM w/ context-aware token-wise $\Lambda_{ca}$}     & $| \V | + d$ & 28.26 &  10.40  & \textbf{11.76} & 12.91 \\
+ a single adaptive $t$     & $| \V | + d + 1$ & 28.61 & 10.37 & \textbf{11.76} & 12.86 \\
+ neighbor-wise adaptive $T$ & $| \V | + d + k$ & 28.69 & 10.39 & 11.79 & 12.89\\
\bottomrule
\end{tabular}
\end{sc}
\end{small}
\end{center}
\vskip -0.1in
\caption{Domain Adaptation performance of the pretrained GPT-2 model (117M) on four language modeling corpora. The best results for each corpus are highlighted in {\bf boldface}. Our approach improves over the baseline $k$NN-LM results (highlighted in \underline{underline}) by 1 $\sim$ 3 perplexity points. Vocab size $| \V |=50257$, the number of retrieved neighbors $k=1024$, and word/sentence embedding size $d=768$. Adaptive parameters are initialized according to Appendix~\ref{sec:app_exp_details}.}
\label{main_table}
\end{table*}

\section{Experiments}

Our experiments aim to understand the following questions:
\begin{enumerate}
    \vspace{-2mm}
    \item What is the utility gain achieved by the \Adapter~over the baseline $k$-Nearest Neighbors Language Model ($k$NN-LM) on various corpora and model sizes (Section~\ref{sec:main_results})?
    \vspace{-2mm}
    \item Which design choice of the \Adapter~is most effective in achieving the utility gain (Section~\ref{sec:main_results})?
    \vspace{-2mm}
    \item How does the \Adapter~perform under real-world limitations, including limited API access and scarce data (Section~\ref{sec:limitation})?
    \vspace{-2mm}
    \item How does the \Adapter~compare to fine-tuning in terms of token efficiency, even though it is proposed as an alternative when fine-tuning is infeasible (Section~\ref{sec:compare_w_finetune})?
\end{enumerate}
% \yang{TODO: list the questions we investigate}

\subsection{Experimental Setup}

\subsubsection{Datasets}
Our evaluation uses the following corpora:

\vspace{-2.5mm}
\paragraph{WikiText-103.} The WikiText-103 benchmark~\cite{merity2016pointer} is a benchmark for evaluating autoregressive language modeling performance. It comprises 103 million tokens from Wikipedia articles in the training set, and 250K tokens in the development and test sets. Our experiments use the original split of the dataset.

\vspace{-2.5mm}
\paragraph{BookCorpus.} BookCorpus~\cite{zhu2015aligning} is a large collection of free novel books written by unpublished authors, which contains 11,038 books (around 74M sentences and 1G words) of 16 different sub-genres (e.g., Romance, Historical, Adventure, etc.). Our experiments preprocess the corpus to sentence level and use three disjoint subsets of 100,000/2,000/2,000 sentences for train/validation/test.

\vspace{-2.5mm}
\paragraph{Enron Email.} The Enron email dataset~\cite{enron2004} is a collection of emails and other documents from the Enron Corporation. The dataset contains over half a million emails, with metadata such as the sender, recipient, and date. We preprocess each email by removing the email header and footer and only keeping the email body. We then use three disjoint subsets of  480,000/5,000/5,000 emails for train/validation/test.

\vspace{-2.5mm}
\paragraph{Amazon Reviews.} The Amazon reviews dataset~\cite{he2016ups} is a large collection of customer reviews and ratings of various products sold on the Amazon website. We preprocess each review into sentence-level and then use three disjoint subsets of  100,000/5,000/5,000 sentences for train/validation/test.

\subsubsection{Models and training details}

We use the pre-trained GPT-2~\cite{radford2019language} small model (117M parameters) as the architecture unless otherwise specified\footnote{Note that the GPT-2 pretraining corpus does not include any of the four corpora mentioned above.}. We use the white-box GPT-2 model rather than the black-box GPT-3 model, because it allows the comparison with model fine-tuning (Section~\ref{sec:compare_w_finetune}). For the $k$NN-LM, we set $k$ (the number of retrieved nearest neighbors) to 1024, and use the $\ell_2$ distance as the distance function for neighbor retrieval. Following~\citet{khandelwal2019generalization}, we use  Faiss~\cite{johnson2019billion} to accelerate the $k$-nearest neighbors search. 

The evaluation reports perplexity on the test set of each corpus. The retrieval datastore is built using the corpus's training set, and the \Adapter~is trained using the corpus's validation set.   
For $k$NN-LM results, we tune hyperparameters on the validation set and report the perplexity on the test set using the best hyperparameters.

\begin{table}[t]
\vskip -0.1in
\setlength{\tabcolsep}{2pt}
\begin{center}
\begin{small}
\begin{sc}
\begin{tabular}{lcccc}
\toprule
 {\bf GPT-2 Model} &  {\bf $\#$ Param.} & {\bf Standard LM} & {\bf $k$NN-LM} & {\bf \Adapter~}\\
\midrule
Small & 117M & 20.22 & 12.49 & 10.35 \\
Medium & 355M & 18.33 &  12.35  &  10.00 \\
Large & 774M &  16.51  & 12.03 &  9.42 \\
\bottomrule
\end{tabular}
\end{sc}
\end{small}
\end{center}
\vskip -0.1in
\caption{Domain Adaptation performance of the pretrained GPT-2 model of different sizes on BookCorpus. \Adapter~gives larger improvement over $k$NN-LM for larger models.}
\label{model_size}
\end{table}

\subsection{Main Results}
\label{sec:main_results}

\paragraph{Effective domain adaptation on various benchmarks.} We compare the evaluation perplexity of $k$NN-LM with \Adapter~on different corpora (see Table~\ref{main_table}). For all corpora, the best results by \Adapter~give an improvement of 1 $\sim$ 3 perplexities over the baseline $k$NN-LM. The training process of the \Adapter~only incurs a slight extra computational cost (fewer than 0.3 GPU hours on a single NVIDIA 2080Ti GPU card).

\begin{figure}[t]
\vskip -0.1in
\begin{center}
\includegraphics[width=\columnwidth]{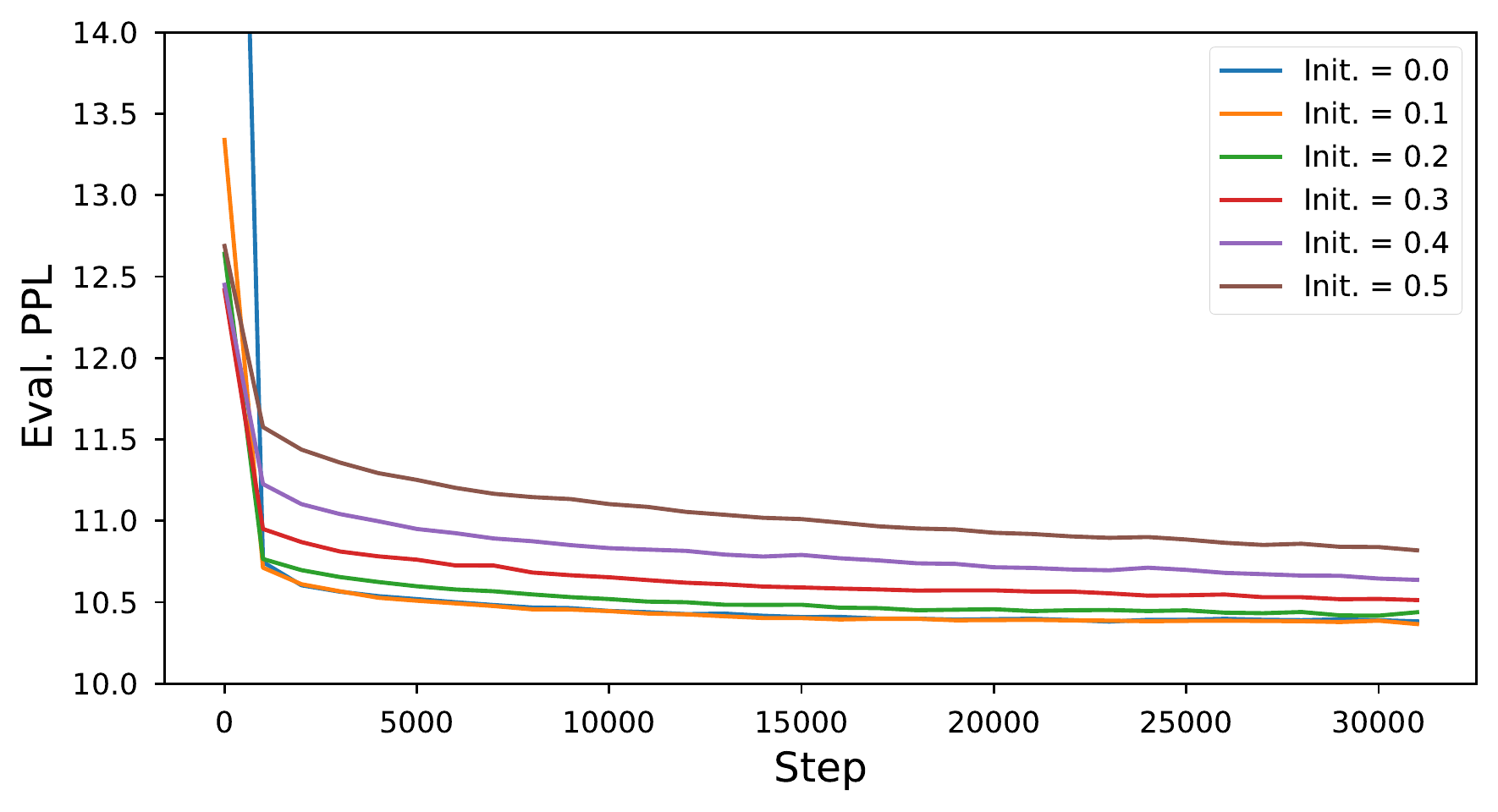}
\end{center}
\vskip -0.2in
\caption{Evaluation perplexity with different initialized values for the adaptive token-wise $\Lambda$. Initializing all entries of $\Lambda$ to 0.1 achieves the best final perplexity.}
\label{compare_init}
\end{figure}

\vspace{-2mm}
\paragraph{Token-wise $\lambda$ gives the most performance improvement.} We also evaluate different combinations of design choices for adaptive interpolation coefficient (see Section~\ref{sec:adapt_lmd}) and adaptive temperature (see Section~\ref{sec:adapt_temp}). As shown in Table~\ref{main_table}), training a token-wise $\Lambda$, as opposed to using a fixed $\lambda$ for all tokens, can lead to improved performance because it allows for more fine-grained control over the weighting of different tokens in the model. For some benchmarks such as Enron Email, the \Adapter~achieves even better performance when the token-wise interpolation coefficient  can be adjusted based on the specific context (i.e., $\Lambda_{ca}$). However, when it comes to the design choices of the adaptive temperature, we find that it only slightly improves performance. Moreover, there is no significant difference between using a single adaptive $t$ for all neighbors or using a customized $T$ for each neighbor. Therefore, for a better utility-efficiency trade-off, our later evaluations use the \Adapter~with adaptive token-wise $\Lambda$ + a single adaptive $t$.

\vspace{-2mm}
\paragraph{Initialization value of $\Lambda$ is important.}  We also find it important to use a good initialization value for the adaptive token-wise $\Lambda$. As shown in Figure~\ref{compare_init}, although initializing all entries in the adaptive $\Lambda$ to 0.1 gives worse results at step$~=0$ (i.e., baseline $k$NN-LM 
 performance) than initializing it to 0.2, it ultimately achieves a better final perplexity. A possible explanation is that uncommon tokens, which are not present in the training of the \Adapter~and thus not updated, perform better with a smaller value of interpolation coefficient. Our further examination in Section~\ref{sec:analysis_freq} validates this explanation.

 Appendix~\ref{sec:app_more_exp} provides more information on the effects of initialization on different corpora. It appears that a small initialized value (such as 0.1 or 0.2) consistently leads to the best final performance of the \Adapter.

\vspace{-2mm}
\paragraph{Larger models benefit more from the \Adapter.}  We also evaluate the performance of \Adapter~when incorporated with GPT-2 models of different sizes. Table~\ref{model_size} shows that the improvement introduced by the \Adapter~is more significant for larger models. This is likely due to the fact that larger models possess higher dimensional sentence embeddings, providing more space for the \Adapter~to make adjustments and fine-tune the model's performance. Furthermore, larger models have more capacity to learn and adapt to new tasks, allowing the \Adapter~to have a greater impact on their overall performance. It's also worth noting that larger models often need more computational resources and training data, which is where the advantages of using the \Adapter~are more prominent.

\iffalse
\begin{table}[t]
\setlength{\tabcolsep}{6pt}
\vskip 0.15in
\begin{center}
\begin{small}
\begin{sc}
\begin{tabular}{lcccr}
\toprule
{\bf Access to Prob.} & {\bf Standard LM} & {\bf $k$NN-LM} & {\bf 
 Adapter}\\
\midrule
% Full  &  \\
Top-10  &  75.69  & 79.17& 80.25 \\ 
Top-5  & 70.80 & 74.02 & 75.72 \\  
Top-3  & 66.96 & 69.70 & 72.00 \\  
Top-1  & 58.58 & 59.27 & 61.77\\  
\bottomrule
\end{tabular}
\end{sc}
\end{small}
\end{center}
\vskip -0.1in
\caption{Domain Adaptation performance measured by top-$q$ accuracy ($\%$) of the pretrained
GPT-2 model on BookCorpus, under different access to the probability distribution over the vocabulary. Having limited access to the probability distribution  results in worse standard LM and $k$NN-LM accuracy, however, using \Adapter~can make up for the accuracy loss. }
\label{access}
\end{table}
\fi

\begin{table}[t]
\setlength{\tabcolsep}{5pt}
\vskip 0.05in
\begin{center}
\begin{small}
\begin{sc}
\begin{tabular}{lcccr}
\toprule
{\bf Access to Prob.} & {\bf Standard LM} & {\bf $k$NN-LM} & {\bf 
 \Adapter}\\
\midrule
Full  & 20.22 & 12.49 & 10.35 \\
Top-10  & 63.18 & 18.96 & 16.00 & \\ 
Top-5  & 85.83 & 20.70 & 17.39 \\  
Top-3  & 113.03 & 22.62 & 18.95 \\  
Top-1  & 210.49 & 27.17 & 22.34\\  
\bottomrule
\end{tabular}
\end{sc}
\end{small}
\end{center}
\vskip -0.1in
\caption{Domain Adaptation performance of the pretrained
GPT-2 model on BookCorpus, under different access to the probability distribution over the vocabulary. Having limited access to the probability distribution  results in worse standard LM and $k$NN-LM perplexity, however, using \Adapter~can greatly make up for the utility loss.}
\label{access}
\end{table}

\begin{table}[t]
\vskip 0.05in
\setlength{\tabcolsep}{4.5pt}
\begin{center}
\begin{small}
\begin{sc}
\begin{tabular}{lcccr}
\toprule
{\bf Datastore} & {\bf Standard LM} & {\bf $k$NN-LM} & {\bf \Adapter} \\
\midrule
WikiText-103 & 20.22 & 13.96 & 12.12 \\
Enron & 20.22 & 18.59 & 15.76 \\
Amazon Reviews & 20.22 & 17.18 & 15.72 \\
\bottomrule
\end{tabular}
\end{sc}
\end{small}
\end{center}
\vskip -0.1in
\caption{Domain Adaptation performance of the pretrained GPT-2 model on BookCorpus, with the datastore built from a different corpora. Our \Adapter~can still achieve significant improvement over baseline $k$NN-LM even if examples in the datastore are not drawn from the target domain.}
\label{limited_data}
\end{table}

\subsection{Domain Adaptation with Limitations}
\label{sec:limitation}

\vspace{-2mm}
\paragraph{Limited API access to $p_{\rm LM}$.} The previous results have assumed access to the full probability distribution over all possible tokens in the vocabulary (i.e., $p_{\rm LM}$). However, due to the size of the vocabulary and computational constraints, the language model API may only output the probability of the top few most likely tokens, rather than the full $p_{\rm LM}$\footnote{For instance, the maximum value for the number of top-most likely tokens in \href{https://beta.openai.com/docs/api-reference/completions/create\#completions/create-logprobs}{OpenAI's API for text completion} is only 5.}. Therefore, we also evaluate the \Adapter's performance under limited access to $p_{\rm LM}$, namely only having access to the top-$q$ ($q < |\V|$) tokens and their probabilities.

As shown in Table~\ref{access}, limiting access to $p_{\rm LM}$  significantly harms the perplexity\footnote{To compute perplexity, we set the probability of the unavailable tokens to $(1-p_{\rm LM, \sum_q})/(|\V|- q)$, where $p_{\rm LM, \sum_q}$ is the sum of top-$q$ tokens' probabilities. It is worth mentioning that using a different way to assign probabilities to unavailable tokens may result in an improved perplexity for all three models (i.e., standard LM, $k$NN-LM, and \Adapter).} of both standard LM and $k$NN-LM. For example, when only the top-5 tokens and their probabilities are available, the perplexity of the standard LM decreases from 20.22 to 85.83 and that of the $k$NN-LM decreases from 12.49 to 20.70. However, using the interpolation strategy learned by the \Adapter~can help  mitigate this utility loss, resulting in an improved perplexity of 17.39. It is also worth noting that the \Adapter's improvement over $k$NN-LM becomes more significant as access to $p_{\rm LM}$ is further limited, as seen in the difference in perplexities between full access (around 2) and only having the top-1 probability (around 5).

\vspace{-2mm}
\paragraph{Scarce data.} We also consider the limitation due to scare data: in real-world situations where there is very limited data available from the target domain, it can even be challenging to build a sufficiently large datastore (for the $k$NN-LM to operate on). However, as shown in Table~\ref{limited_data}, our \Adapter~method can still achieve significant improvements on the target domain (BookBorpus) even when the datastore is built from a {\it different} domain (WikiText, Enron Email, Amazon Reviews). This suggests that the \Adapter~learns to interpolate the output of the language model with retrieval results in a way that is specific to the target domain, regardless of where the datastore originates from. This implies that the \Adapter~is highly versatile and can be applied to a wide range of scenarios, even when data from the target domain is very scarce.

\begin{figure}[t]
% \vskip 0.1in
\begin{center}
\includegraphics[width=\columnwidth]{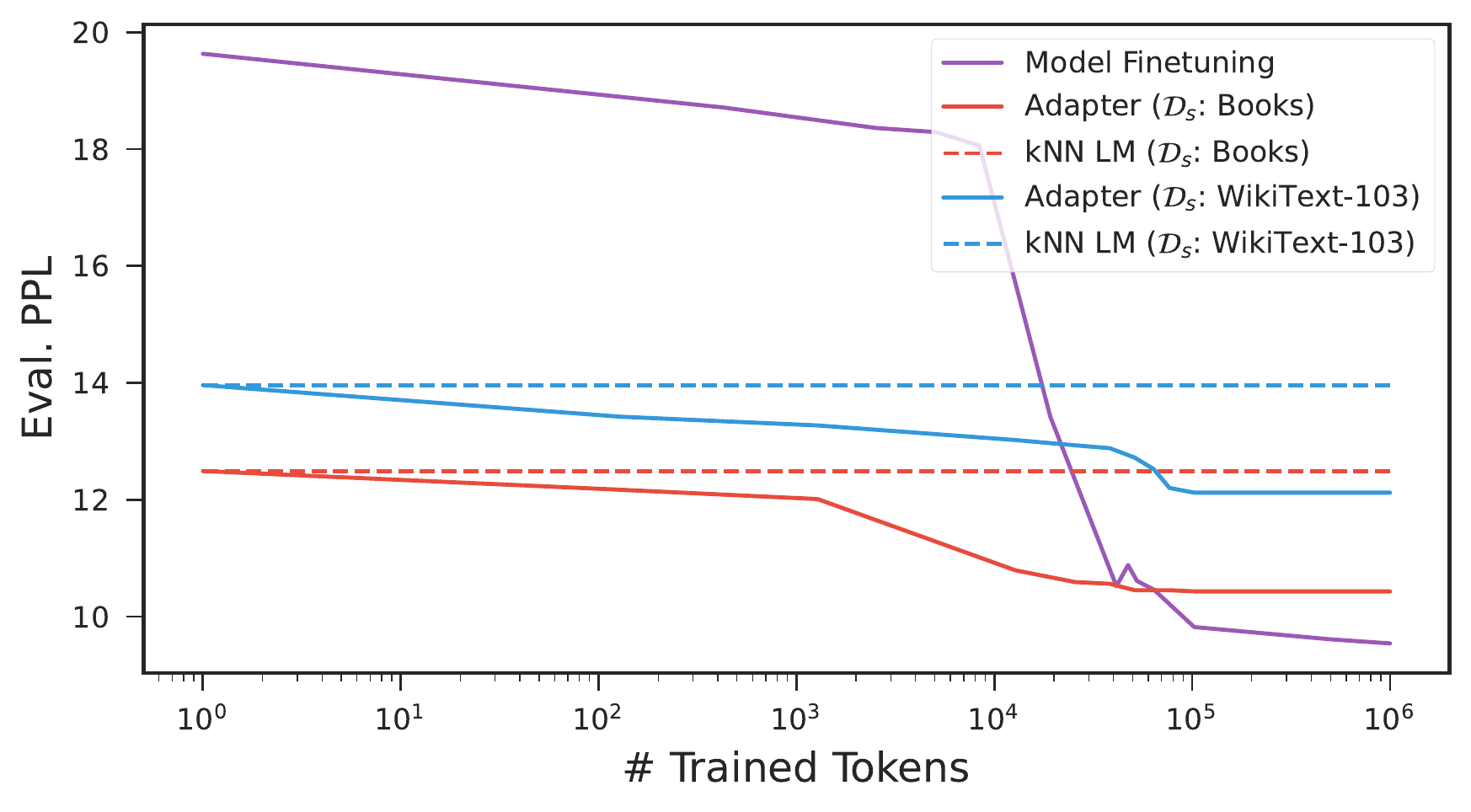}
\end{center}
\vskip -0.2in
\caption{Evaluation perplexity on BooksCorpus for model fine-tuning and \Adapter, with different number of trained tokens.}
\label{compare_w_finetune}
\end{figure}

\subsection{Comparison with Model Fine-tuning}
\label{sec:compare_w_finetune}

We further note \Adapter~may not only serve as a compromise solution when fine-tuning a language model is not possible due to black-box access, but it also has the potential to be a viable alternative to fine-tuning in few-shot settings.

Figure~\ref{compare_w_finetune} reports the evaluation perplexity on BooksCorpus for model fine-tuning and \Adapter. As shown, while fine-tuning may perform better when a large number of tokens are trained (over $10^6$), the \Adapter~can achieve comparable or even superior results when fewer tokens are trained (less than $10^4$). More important, \Adapter (with at most $50$k parameters)~is much cheaper to train compared to fine-tuning the whole model (with 117M parameters).
\section{Analysis of Learned Parameters}
\label{sec:analysis}

In this section, we analyze the adaptively learned parameters in the \Adapter, with the goal of gaining a deeper understanding of the role and behavior of these parameters in $k$-Nearest Neighbor Language Models. We hope this preliminary analysis can help guide future research and development efforts in this area, and lead to the design of more effective and efficient retrieval-augmented language models.

Specifically, we focus on the learned adaptive token-wise $\Lambda$, as it has been shown to contribute most of the performance gain among all design choices (see Section~\ref{sec:main_results}). Our analysis encompasses two perspectives: token frequency (Section~\ref{sec:analysis_freq}) and token's syntactic function (Section~\ref{sec:analysis_function}).

\subsection{Token Frequency}
\label{sec:analysis_freq}

We first analyze the relationship between token frequency and the learned $\lambda$, using Spearman correlation.
We investigate both 1) the token frequency in the pretraining corpus of the language model (GPT-2 in our case), and 2) the token frequency in the datastore of $k$NN-LM.

\begin{table}[t]
% \vskip 0.1in
\setlength{\tabcolsep}{2.5pt}
\begin{center}
\begin{small}
\begin{sc}
\begin{tabular}{lccccc}
\toprule
 & {\bf WikiText-103} & {\bf BookCorpus} &  {\bf Enron} & {\bf Amazon}\\
\midrule
\multicolumn{5}{c}{Learned $\lambda$ v.s. Frequency in Pretraining Corpus} \\
\midrule
Corr.  &	-0.27 & 	-0.21	& -0.29	 & -0.19 \\
p-value	& 0 &	4.48E-122	& 3.59E-177	& 5.15E-94 \\
\midrule
\multicolumn{5}{c}{Learned $\lambda$ v.s. Frequency in Datastore} \\
\midrule
Corr.    &	-0.39	& -0.43	& -0.35	 & -0.51 \\
p-value	    	& 0 &	0	& 2.32E-260	& 0 \\
\midrule
\multicolumn{5}{c}{Frequency in Pretraining Corpus v.s.  in Datastore} \\
\midrule
Corr.  	 & 0.45	  & 0.43  &	0.48  &	0.46 \\
p-value  	& 0 & 0 & 0 & 0 \\
\bottomrule
\end{tabular}
\end{sc}
\end{small}
\end{center}
\vskip -0.1in
\caption{Pair-wise Spearman correlation and p-values for 1) learned $\lambda$, 2) token frequency in the pretraining corpus of GPT-2, and 3) token frequency in the datastore. Both frequencies are negatively correlated with the learned $\lambda$. }
\label{analysis_freq}
\end{table}

As shown in Table~\ref{analysis_freq}, the token's learned $\lambda$ has a negative correlation with the token's frequency in the pretraining corpus. 
This indicates that tokens that have a higher frequency in the pretraining corpus are assigned smaller $\lambda$ values during adaptation to the target domain. This suggests that common tokens, which were heavily emphasized during the language model's pretraining, receive less weight during adaptation, potentially because the model has already demonstrated proficiency with these tokens. In contrast, tokens with low frequency in the language model pretraining corpus are assigned higher weights. It implies that rare tokens benefit more from the $k$NN distribution, aligning with the findings of \citet{min2022nonparametric,zhong2022training} that suggests that $k$NN is better at handling rare patterns.

Regarding the token frequency in the datastore, it was expected that most frequent tokens would result in a better $p_{\rm kNN}$ and subsequently a larger adaptive $\lambda$. However, we find that there is a negative correlation between the learned $\lambda$ and the token frequency in the datastore. This is due to the fact that the frequency in datastore is highly correlated (Spearman correlation $> 0.4$) with the frequency in the pretraining corpus. As a result, tokens that are common in the datastore also likely to be common during pretraining, and thus are assigned a smaller $\lambda$ in the adaptation process.

\subsection{Syntactic Function}
\label{sec:analysis_function}

Another lens to understand the performance gain of the \Adapter~is through its effect on tokens of different syntactic functions. To study this, we categorize tokens by their part-of-speech (POS), including noun, pronoun, adjective, verb, adverb, preposition, conjunction, and interjection\footnote{We use NLTK's \href{https://www.nltk.org/book/ch05.html}{part-of-speech tagging API} to detect the POS of a token.}. We then investigate the group-wised learned $\lambda$'s. In our analysis, we exclude conjunction and interjection as their corresponding tokens are fewer than 10.

\begin{figure}[t]
% \vskip 0.1in
\begin{center}
\includegraphics[width=\columnwidth]{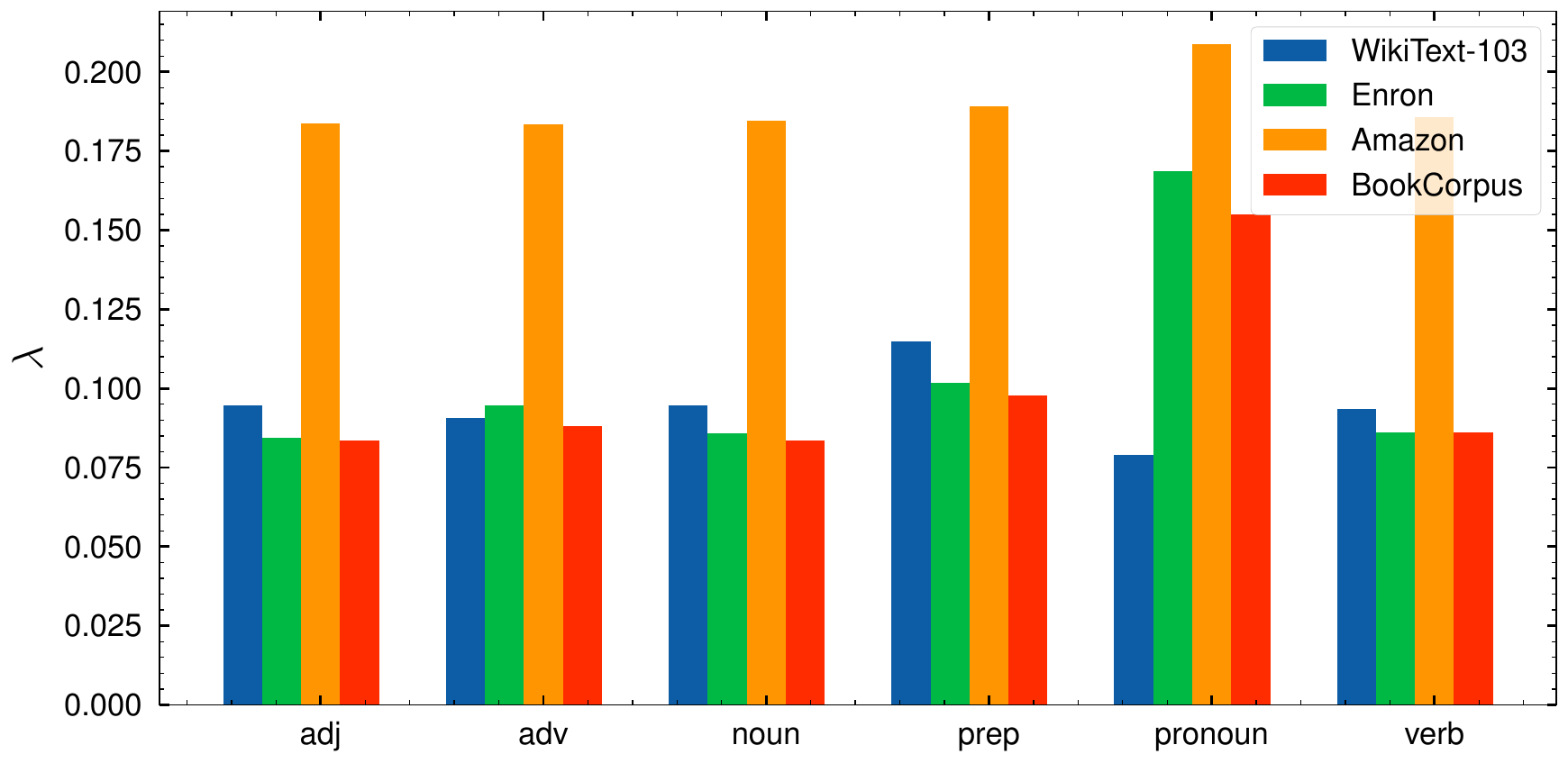}
\end{center}
\vskip -0.2in
\caption{Learned $\lambda$ broken down by token's syntactic functions. In Enron, Amazon and BookCorpus, the learned $\lambda$ for pronoun tokens are highest among the four groups, while in Wikitext-103, the learned $\lambda$ for preposition tokens is highest.}
% \yang{TODO: some text to summarize the findings}} 
\label{fig:function}
\end{figure}

Figure~\ref{fig:function} shows that learned $\lambda$ broken down by tokens' syntactic functions across the four domains. We observe that overall the learned $\lambda$ for Amazon is significantly higher than for the other three domains. This is because the initialization value of $\lambda$ is set to 0.2 for Amazon, whereas it is set to 0.1 for other corpora (see Appendix~\ref{sec:app_exp_details}).

In addition, in Enron, Amazon, and BookCorpus domain, tokens belonging to pronoun group are assigned the highest $\lambda$ among the six groups' tokens, while in Wikitext-103, pronoun tokens have the lowest $\lambda$ among all token groups. This is likely because the Wikitext-103 datastore has significantly fewer pronoun tokens (0.43\%) as shown in Appedix~\ref{sec:app_more_exp}. 
\section{Related Work}

\subsection{Domain Adaptation for Language Models}
% \vspace{-1mm}
It has been shown that simply continual pre-training on a target domain is an effective method to adapt the models to the downstream tasks of that domain~\cite{gururangan2020don}.
This requires full forward and backward passes of the whole model to update it.
Our method focuses on language modeling tasks and does not continue updating the model on the target domain.
Dynamic evaluation~\cite{pmlr-v80-krause18a} adapts language models during inference using gradient information from previous tokens, which makes orthogonal contributions to our work.
Similarly, FWLs~\cite{clark2022meta} play as a light version of dynamic evaluation and avoid back-propagation through the whole transformer encoder, sharing the same inspiration of our method.

% \vspace{-1mm}
\subsection{Retrieval-augmented Language Models}
% \vspace{-1mm}
Retrieval-augmented language models have been widely studied in recent years. 
These models not only rely on encoder forward running, but also leverage a non-parametric component to incorporate more knowledge during inference. 
There are several approaches for integrating retrieved information, such as integrating it at the input space~\cite{guu2020retrieval,izacard2022few, yasunaga2022retrieval}; injecting latent embeddings into intermediate layers of the models~\cite{borgeaud2021improving,de2021mention}; or using an interpolation technique at the output space to merge two token distributions~\cite{khandelwal2019generalization,yogatama2021adaptive,zhong2022training, knnprompt}.
Our work is built on top of the kNN-LM~\cite{khandelwal2019generalization} model, which fuses retrieved information at the output layer.
$k$NN-LM are easily adapted to other domains without architecture changes or re-training. 
This opens up possibilities to train a lightweight adapter that can enhance the effectiveness of domain adaptation even further.

Recent works have proposed approaches to improve the vanilla $k$NN-LM at the inference time~\cite{he2021efficient,alon2022neuro,drozdov2022you,meng2021gnn,zheng2021adaptive}. 
\citet{alon2022neuro} leverages an automaton built on top of the datastore to improve the retrieval effectiveness; \citet{meng2021gnn} integrates the retrieved tokens with a GNN network.
These works are orthogonal to our contribution.
Our work is highly related to \citet{he2021efficient,drozdov2022you}, where they leverage an adaptive interpolation coefficient to enhance $k$NN-LM; however, although also being adaptive, the coefficients for all token types are the same at one position. 
In our paper, we show that using a token-wise adaptive coefficient is crucial to further improvement. 
Besides, their approaches operate on a fine-tuned language model, while our study focuses on adapting a pretrained language model without fine-tuning.
\vspace{-1mm}
\section{Conclusion}

In this paper, we present \Adapter, a novel method for adapting large language models to new domains when fine-tuning is not feasible due to black-box access to model parameters. Our method builds on the $k$-Nearest Neighbor Language Model ($k$NN-LM), and learns to adaptively interpolate the output of the language model with retrieval results. Experiments on four different domains show that \Adapter~significantly improves over the baseline $k$NN-LM and works exceptionally well in settings with limited access to the language model, or with limited data. We also use the adaptively learned parameters to gain a better understanding of why $k$NN-LM is helpful.

We hope this study can inspire more exploration of the possibility of adapting black-box language models to new domains. We plan to release the datasets used in our experiments to encourage the further improvement of the \Adapter's accuracy and efficiency.

The major limitation of the current work is the use of  GPT-2 in evaluation because it has white-box access and thus allows the comparison with  fine-tuning. In future work, we plan to extend the evaluation of \Adapter~to real-world black-box language models, e.g., GPT-3, and to more tasks, including sentence classification and machine translation.

% \section*{Acknowledgement}

\newpage
\bibliography{references}
\bibliographystyle{icml2023}

%%%%%%%%%%%%%%%%%%%%%%%%%%%%%%%%%%%%%%%%%%%%%%%%%%%%%%%%%%%%%%%%%%%%%%%%%%%%%%%
%%%%%%%%%%%%%%%%%%%%%%%%%%%%%%%%%%%%%%%%%%%%%%%%%%%%%%%%%%%%%%%%%%%%%%%%%%%%%%%
% APPENDIX
%%%%%%%%%%%%%%%%%%%%%%%%%%%%%%%%%%%%%%%%%%%%%%%%%%%%%%%%%%%%%%%%%%%%%%%%%%%%%%%
%%%%%%%%%%%%%%%%%%%%%%%%%%%%%%%%%%%%%%%%%%%%%%%%%%%%%%%%%%%%%%%%%%%%%%%%%%%%%%%
\newpage
\appendix
\onecolumn
\section{Appendix}
\label{app}

\subsection{Experimental details}
\label{sec:app_exp_details}

We train the \Adapter~uses the Stochastic Gradient Descent (SGD) optimizer, with a learning rate of 0.1 and a batch size of 128. We tune the initialization for adaptive parameters on a held-out validation set, and the best parameters are shown as follow.

\begin{table}[h]
\setlength{\tabcolsep}{2pt}
\vskip 0.15in
\begin{center}
\begin{small}
\begin{sc}
\begin{tabular}{lcccccc}
\toprule
{\bf Param } & {\bf WikiText-103} & {\bf BookCorpus} &  {\bf Enron} & {\bf Amazon Reviews} \\
\midrule
A single interpolation coefficient $\lambda$ & 0.25 & 0.25 & 0.25 & 0.25 \\
Token-wise interpolation coefficient  $\Lambda$ & 0.1 & 0.1 & 0.1 & 0.2 \\
Context-aware token-wise interpolation coefficient  $\Lambda_{ca}$ & 0.1 & 0.1 & 0.1 & 0.2 \\
A single temperature $t$ & 1.0 & 1.0 & 1.0 & 1.0 \\
Neighbor-wise temperature $T$ & 1.0 & 1.0 & 1.0 & 1.0\\
\bottomrule
\end{tabular}
\end{sc}
\end{small}
\end{center}
\vskip -0.1in
\caption{Initialization values for different design choices of \Adapter.}
\label{tab:init_values}
\end{table}

\subsection{More Experimental Results}
\label{sec:app_more_exp}

Table~\ref{best_lmd} reports the evaluation perplexity of the \Adapter~with different initialized values.  It appears that a small
initialized value (such as 0.1 or 0.2) consistently leads to the best final performance.

\begin{table}[h]
\label{ablation_study}
% \vskip 0.15in
\begin{center}
\begin{small}
\begin{sc}
\begin{tabular}{lccccc}
\toprule
{\bf Init. } & {\bf WikiText-103} & {\bf BookCorpus} &  {\bf Enron} & {\bf Amazon Reviews} \\
\midrule
0.0  & 28.66 & 10.38 & 11.86 & 13.07 \\
0.1  & 28.25 & 10.36 & 11.75 & 12.93  \\
0.2  & 28.46 & 10.44 & 11.89 & 12.88 \\
0.3  & 29.06 & 10.51 & 12.13 & 12.89 \\
0.4  & 29.51 & 10.64 & 12.47 & 12.93 \\
0.5  & 29.99 & 10.82 & 12.93 & 13.04 \\
\bottomrule
\end{tabular}
\end{sc}
\end{small}
\end{center}
\vskip -0.1in
\caption{Evaluation perplexity with different initialized values for the adaptive $\lambda$.}
\label{best_lmd}
\end{table}

Figure~\ref{fig:function} presents the percentage of token frequency broken down by syntactic function in each domain's datastore.

\begin{figure}[h]
% \vskip 0.1in
\begin{center}
\includegraphics[width=0.6\columnwidth]{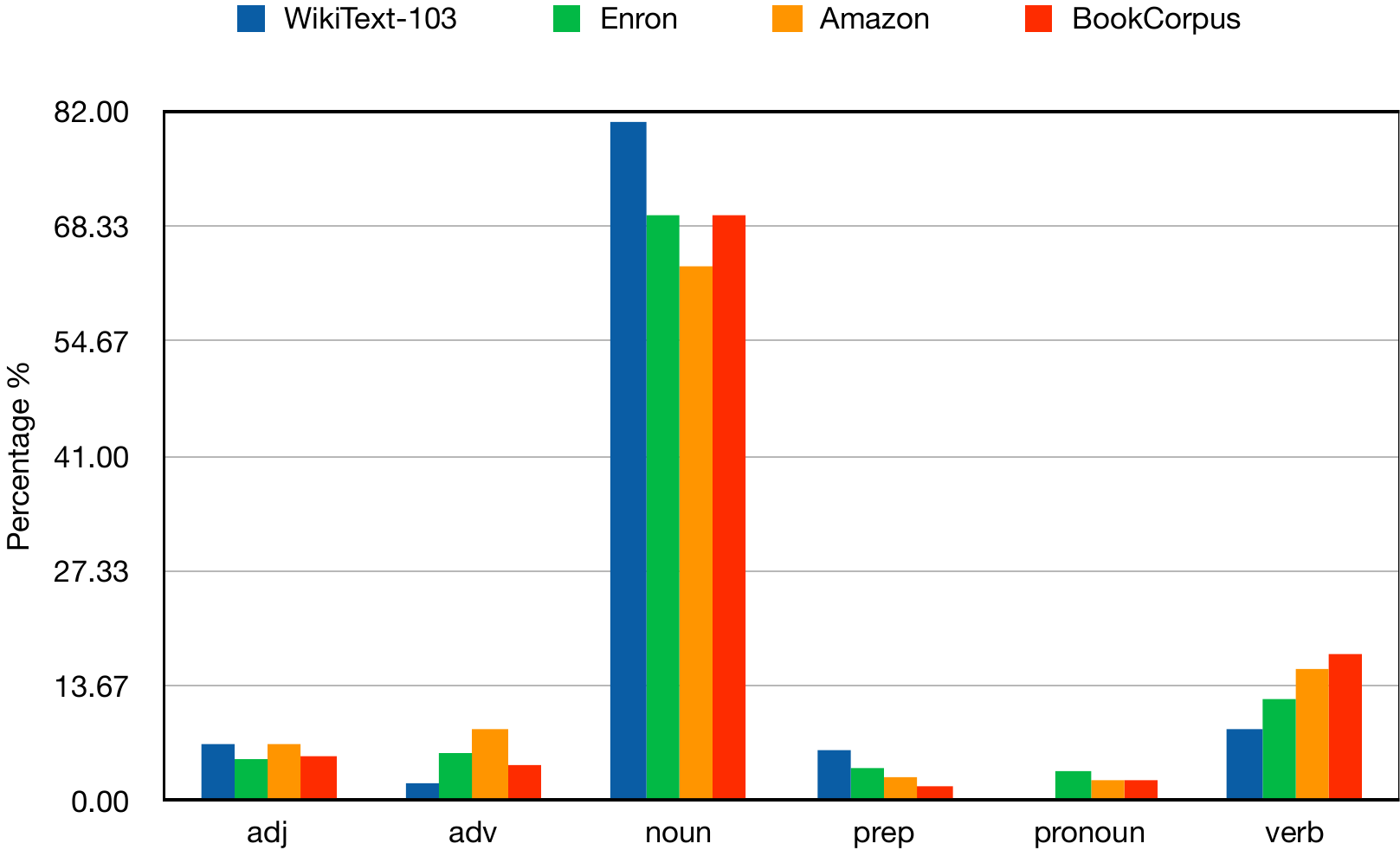}
\end{center}
\vskip -0.2in
\caption{Percentage of frequency of each token's syntactic function in datastore.} 
\label{fig:token_type_percentage}
\end{figure}

%%%%%%%%%%%%%%%%%%%%%%%%%%%%%%%%%%%%%%%%%%%%%%%%%%%%%%%%%%%%%%%%%%%%%%%%%%%%%%%
%%%%%%%%%%%%%%%%%%%%%%%%%%%%%%%%%%%%%%%%%%%%%%%%%%%%%%%%%%%%%%%%%%%%%%%%%%%%%%%

\end{document}